\documentclass{ecai} 

\usepackage{latexsym}
\usepackage{amssymb}
\usepackage{amsmath}
\usepackage{amsthm}
\usepackage{booktabs}
\usepackage{enumitem}
\usepackage{graphicx}
\usepackage{color}

\usepackage{algorithmic}
\usepackage{algorithm}
\usepackage{bm}
\usepackage{amsfonts, pxfonts}
\usepackage{multirow}
\usepackage{makecell}
\usepackage{listings}
\usepackage{url}
\usepackage{tabularx}
\usepackage[title]{appendix}
\usepackage{xr}

\newcommand{\BibTeX}{B\kern-.05em{\sc i\kern-.025em b}\kern-.08em\TeX}

\makeatletter
\newcommand*{\addFileDependency}[1]{
  \typeout{(#1)}
  \@addtofilelist{#1}
  \IfFileExists{#1}{}{\typeout{No file #1.}}
}
\makeatother

\newcommand*{\myexternaldocument}[1]{%
    \externaldocument{#1}%
    \addFileDependency{#1.tex}%
    \addFileDependency{#1.aux}%
}

\myexternaldocument{appendix}

\begin{document}


\begin{frontmatter}


\paperid{m915} 


\title{CLLMFS: A Contrastive Learning enhanced Large Language Model Framework for Few-Shot Named Entity Recognition}


\author[A]{\fnms{Yafeng}~\snm{Zhang}\orcid{0000-0001-5619-1721}\thanks{Corresponding Author. Email: yfzhang40@iflytek.com.}\footnote{Equal contribution.}}
\author[B]{\fnms{Zilan}~\snm{Yu}\orcid{0000-0002-9460-0984}\thanks{Corresponding Author. Email: yuzl22@mails.tsinghua.edu.cn.}\footnote{Equal contribution.}}
\author[A]{\fnms{Yuang}~\snm{Huang}\orcid{0009-0004-9084-6807}}
\author[D]{\fnms{Jing}~\snm{Tang}\orcid{0000-0002-9430-9660}} 

\address[A]{iFLYTEK Co., Ltd.}
\address[B]{Tsinghua University}
\address[C]{Huazhong University of Science and Technology}


\begin{abstract}
Few-shot Named Entity Recognition (NER), the task of identifying named entities with only a limited amount of labeled data, has gained increasing significance in natural language processing. While existing methodologies have shown some effectiveness, such as enriching label semantics through various prompting modes or employing metric learning techniques, their performance exhibits limited robustness across diverse domains due to the lack of rich knowledge in their pre-trained models. To address this issue, we propose CLLMFS, a Contrastive Learning enhanced Large Language Model (LLM) Framework for Few-Shot Named Entity Recognition, achieving promising results with limited training data. Considering the impact of LLM's internal representations on downstream tasks, CLLMFS integrates Low-Rank Adaptation (LoRA) and contrastive learning mechanisms specifically tailored for few-shot NER. By enhancing the model's internal representations, CLLMFS effectively improves both entity boundary awareness ability and entity recognition accuracy. Our method has achieved state-of-the-art performance improvements on F1-score ranging from 2.58\% to 97.74\% over existing best-performing methods across several recognized benchmarks. Furthermore, through cross-domain NER experiments conducted on multiple datasets, we have further validated the robust generalization capability of our method. Our code will be released in the near future.

\end{abstract}

\end{frontmatter}


\section{Introduction}

Named Entity Recognition (NER) is pivotal for identifying and categorizing named entities within unstructured text across various domains, such as Location \citep{tjong-kim-sang-de-meulder-2003-introduction}, Private Health Information \citep{stubbs2015identifying} and Event \citep{weischedel2013ontonotes}. However, developing accurate NER models demands substantial amounts of domain-specific annotated data, which are often scarce and costly to procure \citep{huang2021few}. This has led to the demand for Few-Shot Named Entity Recognition (FS-NER), which aims to learn from limited labeled examples to address entity tagging challenges under low-resource conditions \citep{chen-etal-2023-prompt}. 

Early FS-NER methods often employ neural networks with conventional supervised learning, which may lead to overfitting due to the large number of parameters to optimize \citep{bejani2021systematic}. To mitigate this, cross-domain NER approaches have been employed, where models learn semantic features from base classes and adapt them to novel classes \citep{min2023recent}. Despite this, these methods may still exhibit suboptimal generalization in novel domains \citep{dodge2020finetuning}. To address these limitations, contrastive learning has been introduced, utilizing Gaussian distributions to optimize the distributional distance between tokens in sentences \citep{container}. 

With the rapid development of Large Language Models (LLMs), models like GPT-3 demonstrate few-shot capabilities through prompt-based construction, achieving satisfactory results \citep{wang2023gptner}. LLAMA 2 \citep{llama2} emerges as a superior choice in low-resource settings due to its accessibility and adeptness across various natural language processing (NLP) tasks. However, deploying of LLMs such as ChatGLM, GPT-3, ChatGPT, GPT-4, LLAMA, and LLAMA 2 \citep{du2021glm,brown2020language, ouyang2022training,bubeck2023sparks,touvron2023llama,touvron2023llama2} for FS-NER poses challenges, given their extensive parameter sizes and the need for substantial amounts of high-quality supervised fine-tuning (SFT) data, leading to high costs in training and data acquisition. To address this, Parameter-Efficient Fine-Tuning (PEFT) techniques like Low-Rank Adaptation (LoRA) \citep{lora} have been proposed to enhance model performance on new tasks while minimizing fine-tuning parameters and computational complexity.

In this research, we present an innovative method, CLLMFS, to tackle the NER task under low-resource conditions. Our approach leverages LLMs to effectively address the challenge of limited labeled data by exploiting their pre-trained knowledge. We fine-tune LLMs using supervised learning to adapt them to our specific NER task, resulting in improved performance compared to recent benchmarks. To further reduce the trainable parameters, we employ LoRA techniques, enabling effective fine-tuning with limited training samples. Additionally, we introduce contrastive learning to our framework, enriching the LLM-based method for few-shot NER tasks. This framework significantly improves the boundary awareness of the LLM and enhances its ability to accurately extract named entities by refining internal embedding representations. Furthermore, we enhance the model's robustness by introducing noise to construct positive example pairs during training. Our approach achieves state-of-the-art results across multiple datasets, demonstrating its effectiveness and versatility in handling NER tasks under low-resource settings.

In summary, our contributions are as follows:
\begin{itemize}
    \item We advance the use of LLMs for few-shot NER by integrating them with LoRA for supervised fine-tuning, achieving the-state-of-art performance across multiple datasets with limited labeled data.
    \item We propose a framework that incorporates contrastive learning to improve the boundary awareness and accuracy of entity extraction, enhancing model robustness by constructing positive example pairs with noised embedding. 
    \item Our approach showcases robust transfer capabilities, significantly enhancing the F1-score, ranging from 2.58\% to 97.74\% in the INTRA setting, and from 44.36\% to 160.00\% in the INTER setting, surpassing state-of-the-art methods across various datasets.
\end{itemize}


\section{Related Works}
\label{sec:related works}

\subsection{Few-Shot NER}
Few-shot learning (FS-NER) enhances model performance with limited labeled data. Data-enhancement methods augment small labeled datasets with additional data sources, but unreliable examples can affect precision \citep{yang2020simple}. Manner uses a Variational Autoencoder for an external memory module, but faces challenges in memory optimization and cross-domain generalization \citep{fang2023manner}. CONTaiNER employs contrastive learning to optimize token distribution, improving adaptability to new domains \citep{container}, but large source-target domain divergence can be problematic.

\subsection{Meta Learning}
Meta Learning offers new approaches for few-shot learning. Metric-based methods like Matching Networks \citep{Vinyals2016MatchingNF} and Prototypical Networks \citep{Snell2017PrototypicalNF} calculate similarities to learn prototypical representations for target classes. ProtoBERT uses a pre-computable BERT encoder for effective entity prediction \citep{zhang2023less}. ProML introduces multiple prompt schemas with weighted averages for enriched label semantics \citep{chen-etal-2023-prompt}, achieving promising results across various settings. However, generated prototypes may lack precision, due to limited labeled data for various entity types in the support set.

\subsection{In-context learning} 
Large-scale pre-trained LLMs, like GPT-3 \citep{brown2020language}, have advanced in-context learning, applied in tasks like question answering and NER without additional training data \citep{akyurek2022learning}. Recent NLP research explores prompt-based methods for FS-NER, relying on prompts to predict labels. However, these methods primarily rely on prompts to predict labels using classification heads, rather than employing data-enhancement or metric learning techniques. Prompt-based NER uses language models to generate entity predictions based on context and instructions. However, these methods face limitations in prompt quality and design.


\section{Methodology}
\label{sec:methodology}

The internal representations of language models play a pivotal role in shaping the performance of downstream tasks. In this paper, we introduce a novel model, CLLMFS, based on large pre-trained language models. As depicted in Figure~\ref{fig1}, our model undergoes supervised fine-tuning in source domains under the \textit{N-way K-shot} scenario, enabling it to adapt to target domains effectively. Ultimately, our model integrates LoRA and contrastive learning loss techniques, specifically customized for the NER task.

\subsection{Task definition}
\label{ssec:ner-setting}

Given a sequence of \textit{n} tokens \(\{x_1, x_2, \dots, x_n\}\) and corresponding tag labels \(\{y_1, y_2, \dots, y_n\}\), the primary objective of NER is to associate each token \(x_i\) with its corresponding tag label \(y_i\). In Few-shot NER, a model undergoes training in a low-resource source domain with a tag-set denoted as \(\{C^s_{i}\}\). Subsequently, it is tested in a target domain that employs a distinct tag-set, denoted as \(\{C^d_{j}\}\), where i and j represent indices for different tags. Since \(\{C^s_{i}\} \cap \{C^d_{j}\} = \varnothing\), the model faces the formidable challenge of generalizing to previously unseen test tags. In an \textit{N-way K-shot} scenario, the source domain comprises \textit{N} distinct entity types, denoted as \(|\{C^s_{j})\}| = N\). For each entity type, there are \textit{K} examples in the support set. This setup means that the model is trained with \textit{K} labeled examples for each of the \textit{N} types, enabling it to learn and generalize from a limited number of examples.

\subsection{LLM for entity extraction}
\label{ssec:model}

\begin{figure*}[t]
\begin{center}
\centerline{\includegraphics[width=1\linewidth]{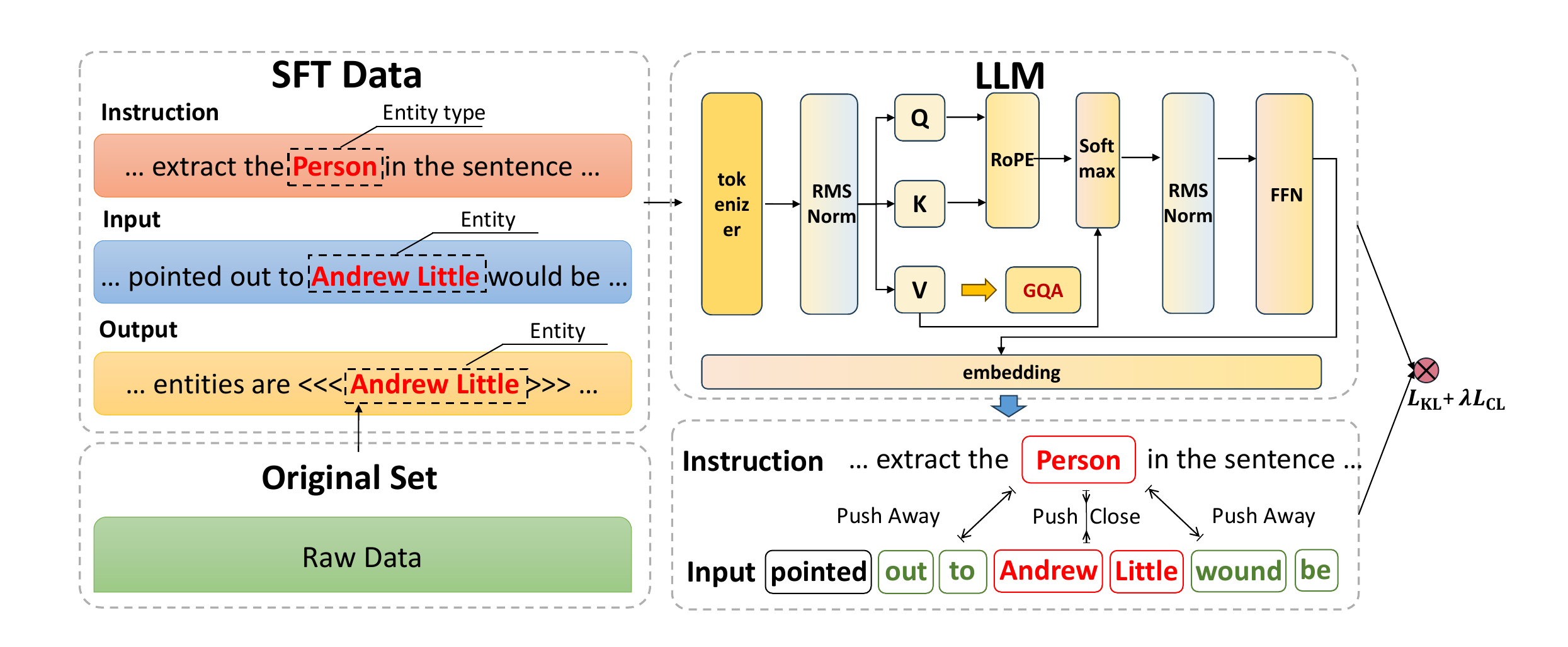}}\vspace{-.3cm}
\caption{The framework overview CLLMFS. The LLM extracts named entities from carefully designed SFT data using decoding strategies, LoRA fine-tuning leveraging LLM's attention mechanisms such as QKV computations, constructing positive and negative samples for contrastive learning, and creating adversarial embedding samples.}
\label{fig1}\vspace{-.3cm}
\end{center}
\end{figure*}

Our method for entity extraction tasks is based on LLAMA 2, referred to as \(\mathrm{LLM}\). Figure~\ref{fig1} illustrates the pivotal components of our model. We utilize the 7-billion parameter model of LLAMA 2, balancing effectiveness and inference speed. Our approach achieves excellent results in entity extraction tasks with only a small amount of training samples.

LLAMA 2's architecture closely resembles the standard Transformer Decoder, primarily consisting of 32 Transformer Blocks. Each block includes the following core components:

\begin{itemize}
    \item \textbf{RMSNorm} \citep{Zhang2019RootMS}: Normalizes the activation outputs of network layers, ensuring uniform scaling, accelerating training, and enhancing model stability.
    \item \textbf{SwiGLU} \citep{Shazeer2020GLUVI}: Adds non-linearity to the model by transforming input values through the Swish activation function.
    \item \textbf{RoPE} \citep{Su2021RoFormerET}: A novel positional encoding strategy that encodes positional information through rotation operations.
    \item \textbf{GQA} \citep{ainslie-etal-2023-gqa}: Divides query heads into G groups, with each head maintaining its own query parameters and each group sharing a key and value matrix, simplifying calculations and improving the efficiency of attention computation in large models.
\end{itemize}

By leveraging these components, our method effectively extracts entities with high accuracy and efficiency.

\subsection{Model Training}
\subsubsection{Supervised fine-tuning with LoRA}
Supervised fine-tuning (SFT) refers to the process of adjusting a pre-trained LLM using labeled data to better adapt it to a specific task. During SFT, weights of the model are adjusted based on the discrepancies with the true labels, aiming to enhance precision and task adaptation.

Each sample in SFT typically consists of three parts: instruction (i.e., prompt), input, and output. For instance, for the entity type "Person" (other entity types are provided in the Appendix A):

\texttt{
\{
  "instruction": "Please extract the Person in the sentence given below, the entity of person refers to the entity that represents the identity or role of a specific person in the input sentence.",
  "input": "True , but I imagine it would be a lot lower and as I pointed out to Andrew Little would be cheaper than [ eliminating fees .",
  "output": "<im\_start> I can extract entities for you, the extracted entities are \text{<}\text{<}\text{<} Andrew Little \text{>}\text{>}\text{>} <im\_end>"
\}
}

The design concept behind constructing SFT data involves using the instruction to define the entity extraction task and the types of entities to be extracted, guiding the LLM to efficiently perform entity extraction tasks. The input represents the original input of the user, containing the sentences from which entities are to be extracted. For example, Our entity type token in \texttt{instruction} is "Person", and the actual entity in the \texttt{input} sentence is "Andrew Little". The output denotes the output results of the model, with the extracted entities surrounded by start (\texttt{\text{<}\text{<}\text{<}}) and end (\texttt{\text{>}\text{>}\text{>}}) symbols. Additionally, we intentionally devised a specific format for the output of the LLM, starting with \texttt{<im\_start>} and ending with \texttt{<im\_end>}.

Considering the limited amount of the data generated by SFT, full model fine-tuning is not feasible and the issue of overfitting is also serious. We adopt Low-Rank Adaptation (LoRA) \citep{lora} to address these problems. LoRA assumes that weight updates during the adaptation process also have a lower 'intrinsic rank'. For a pre-trained weight matrix \( W_0 \in \mathbb{R}^{d \times k} \), we restrict its update through a low-rank decomposition as follows:

\begin{equation}
W_0 + \Delta W = W_0 + BA
\end{equation}
where \( B \in \mathbb{R}^{d \times r} \), \( A \in \mathbb{R}^{r \times k} \), and the rank \( r << \min(d, k) \). Throughout training, \( W_0 \) remains frozen and does not undergo gradient updates, while \( A \) and \( B \) include trainable parameters. We employ a random Gaussian initialization for \( A \) and set \( B \) to zero, ensuring that \( \Delta W = BA \) is zero at the start of training.

In other words, during the fine-tuning process, the model initializes with pre-trained parameters \(W_0\) and updates them to \(W_0 + \Delta W(\theta)\) by maximizing the conditional language model probability, where \(|\theta| << |W_0|\):

\begin{equation}
\max_{\theta} \sum_{(x,y)\in Z} \sum_{t=1}^{|y|} \log(P_{W_0+\Delta W(\theta)}(y_t|x, y < t))
\end{equation}
where \(Z\) denotes the training dataset comprising input sequences \(x\) and their corresponding target sequences \(y\), and \(|y|\) signifies the length of the target sequence \(y\). \(P_{W_0+\Delta W(\theta)}(y_t | x, y < t)\) represents the probability that the model predicts the \(t\)-th element \(y_t\) of the target sequence, given the input sequence \(x\) and the first \(t\) elements of the target sequence \(y\). LoRA fine-tune only need a subset of parameters, thereby avoiding issues such as excessive resource consumption caused by full fine-tuning.

In principle, LoRA can be applied to any subset of weight matrices in a neural network, thereby reducing the number of trainable parameters. 


\subsubsection{Decoding strategies}
\label{ssec:ds}

Decoding strategies play a pivotal role in text generation tasks. In our approach, we maintain a fixed temperature of 0.01 to control the diversity of generated text. Additionally, we employ the top-k sampling method, where only tokens ranking within the top probability threshold (top\_p) are considered during the token sampling process. This strategy optimizes the greedy approach by sampling from the top-k tokens, allowing tokens with higher scores or probabilities beyond the top threshold to also have a chance of being selected.

Given the nature of our task, which involves information extraction, we implement a constrained generation approach. This ensures that the generated output is constrained to be a subset of the input, restricting the generated content within predefined boundaries. Furthermore, to prevent the model from endlessly generating content, we introduce a custom stop symbol, denoted as \texttt{<im\_end>}, marking the end of the generated sequence. This mechanism effectively halts the generation process after the last eight characters, ensuring controlled and targeted generation. Employing a combination of a low-temperature setting and constrained generation proves instrumental in effectively handling few-shot scenarios.

The internal representations of language models have a significant impact on the performance of downstream tasks. In this paper, we employ contrastive learning loss in low-resource entity extraction tasks to enhance the boundary perception ability of the model and improve its effectiveness.

\subsection{Contrastive learning}

Contrastive learning is a discriminative representation learning method based on the principle of comparison, primarily used for unsupervised (self-supervised) representation learning. The core idea of contrastive learning is to compare samples with positive examples (semantically similar) and negative examples (semantically dissimilar). By designing contrastive losses, it aims to bring representations of semantically similar positive examples closer while pushing representations of semantically dissimilar negative examples further apart. Therefore, the careful selection of positive and negative sample pairs for contrastive learning is crucial.

We propose a specific approach to address the challenge of designing positive and negative samples for contrastive learning in low-resource entity extraction tasks. In the constructed SFT data, the entities to be extracted from the input are designated as positive samples. Additionally, the neighboring entities of the target entities serve as negative samples, emphasizing the significance of capturing the entity boundaries accurately. As shown in fig.~\ref{fig1}, the embedding of "Person" is proximately aligned with that of "Andrew Little", while being intentionally distanced from the embeddings corresponding to "out to" and "wound be". It is essential to avoid over-extracting or under-extracting words, particularly ensuring against over-extracting. 

We define \(\mathcal{T}\) as the set of embeddings of entity types within the instructions, and \(\mathcal{E}\) as the set of embeddings of entities from input sentences. Therefore, we treat the entity type embeddings within the instructions and the embeddings of the entities to be extracted from input sentences as the positive pairs (\textit{i.e.},\( \{(z_{instr}^{t} , z_{in}^{e}) | t \in \mathcal{T}, e \in \mathcal{E} \} \)). Simultaneously, we establish negative pairs (\textit{i.e.},\( \{(z_{instr}^{t} , z_{in}^{n}) | t \in \mathcal{T}, n \in \mathcal{E} \} \)) by considering the entity type embeddings within the instructions and the neighboring entities of the entities to be extracted from input sentences. Formally, we employ the contrastive loss, InfoNCE \citep{infoNCE}, to maximize agreement among positive pairs and minimize it among negative pairs:

\begin{equation}
L_{CL} = \sum_{t \in \mathcal{T}, e \in \mathcal{E}} - \log \frac{ \exp{ (s(z_{instr}^{t}, z_{in}^{e})/ \tau)} }{ \sum_{n \in \mathcal{E}, n \neq e} \exp{( s( z_{instr}^{t}, z_{in}^{n} / \tau )} }
\end{equation}
where \(s(\cdot)\) denotes the similarity between two vectors and is set as the cosine similarity function. \(\tau\), referred to as the Temperature parameter in the softmax function, is a hyper-parameter.

By employing this design for positive and negative samples, the distance between the entity type embeddings within the instructions and the embeddings of the entities to be extracted from input sentences is minimized. This adjustment enables the model to prioritize positive entities more effectively during generation. Simultaneously, it increases the distance between the entity type embeddings within the instructions and the neighboring entities of the entities to be extracted from input sentences. This enhances the model's boundary perception ability, resulting in more precise extraction of entity information.


\subsection{Enhancing Representation Uniformity with Adversarial Samples}

The representations generated by contrastive learning are typically regularized, causing them to concentrate within a hypersphere. Alignment and uniformity refer to two essential characteristics of a good representation space: alignment ensures that representations of semantically similar samples are close together, while uniformity ensures that representations of semantically dissimilar samples are evenly distributed across the hypersphere. Enhancing the uniformity of representation distributions can improve the performance of many tasks, such as recommendation systems.

However, previous research has primarily relied on in-batch negative sampling or random negative sampling from the training data. This approach may introduce sampling bias, leading to the inclusion of inappropriate negative examples (such as false negatives or anisotropic representations) in contrastive learning, potentially compromising the alignment and uniformity of the representation space.

To achieve a more uniformly distributed representation space, we focuses on the embedding space and directly introduces noise into the representations. Inspired by \citet{SimGCL}, we construct adversarial samples through imperceptible perturbations by adding uniformly distributed Gaussian random noise to positive embeddings of entities. While this approach is simple, it can strengthen the positive samples to resist noise, leading to a significant enhancement in the model's robustness against interference. Formally, given a token \(i\) and its embedding \(z_i\) in the \(d\)-dimensional space, we can implement the following representation-level augmentation:

\begin{equation}
z^{'}_i = z_i + \Delta_i^{'}
\end{equation}
where \(\Delta_i^{'}\) is the added noise vectors.

\subsection{Model Optimization}

To train our model effectively for the low-resource entity extraction task, we employ a combined loss function comprising both cross-entropy loss and contrastive learning loss.

The primary objective of cross-entropy loss in Few-shot NER is to ensure that our model learns to correctly associate each token \(x_i\) in the input sequence with its corresponding tag label \(y_i\). This involves minimizing the discrepancy between the predicted tag probabilities and the ground truth labels across the entire sequence. Formally, the cross-entropy loss \(L_{CE}\) is computed as follows:

\begin{equation}
L_{CE} = - \sum_{i} \sum_{c \in \{C^d_{j}\}} y_{i,c} \log(\hat{y}_{i,c})
\end{equation}
where \(y_{i,c}\) represents the ground truth label for token $x_i$ corresponding to tag \(c\) in the target domain, and \(\hat{y}_{i,c}\) represents the predicted probability of token \(x_i\) belonging to tag \(c\).

By minimizing the cross-entropy loss, our model learns to accurately predict the tags associated with each token in the input sequence, thereby improving its performance in the NER task, especially when dealing with previously unseen tags in the target domain.

In addition to cross-entropy loss, we incorporate contrastive learning loss to further enhance the model's performance. The contrastive loss \(L_{CL}\) encourages the model to effectively distinguish between positive pairs (tokens associated with the same entity type) and negative pairs (tokens associated with different entity types). 

Due to the presence of 32 Transformer Blocks (i.e., 32 layers of hidden states) in LLAMA 2, we determined the optimal layer for computing the contrastive loss through empirical testing. Configurations using the 10th, 25th, 26th, 27th, and 30th layers were evaluated, and the 26th layer consistently yielded the best performance. This layer selection closely aligns with the 8:2 golden ratio, providing a balance between the lower and higher layers in the model’s architecture. Therefore, we compute the contrastive loss at the 26th layer to leverage this optimal configuration.

Finally, we leverage a multi-task training strategy to jointly optimize the cross-entropy loss, and the contrastive learning loss. The overall loss function is:

\begin{equation}
L = L_{CE} + \lambda L_{CL}
\end{equation}
where \(\lambda\) serves as a hyperparameter to regulate the impact of contrastive learning and is set to 0.001. This choice is made considering that different losses calculate distinct gradients, with the aim of emphasizing the gradient of the main task.

By jointly optimizing cross-entropy loss and contrastive learning loss, our model learns to effectively classify entities while also capturing semantically meaningful representations, thus enhancing its overall performance in the low-resource entity extraction task.


\section{Experiments}
\label{sec:experments}

\begin{table*}
\caption{Overall Performance Comparison.}
    \centering
    \resizebox{\linewidth}{!}{
    \begin{tabular}{lc|ccccc|c}
        \hline
        & \textbf{Model} & WNUT'17 & GUM & I2B2 & OntoNotes & CoNLL'03 & Avg. \\
        \hline \hline
        \multirow{5}{*}{\makecell{INTRA}} &
        ProtoBERT & 0.2655 & 0.1374 & 0.3433 & 0.3818 & 0.3218 & 0.2900\\
        & NNShot & 0.2305 & 0.0683 & 0.3844 & 0.3454 & 0.3382 & 0.2734 \\
        & ProML & 0.2262 & 0.2336 & 0.5654 & 0.2548 & 0.3424 & 0.3249 \\
        & CONTaiNER & 0.2108 & 0.1328 & 0.3807 & 0.2275 & 0.3199 & 0.2543 \\
        & CLLMFS & \textbf{0.5250} & \textbf{0.3840} & \textbf{0.5800} & \textbf{0.5765} & \textbf{0.5750} & \textbf{0.5281} \\
        \hline
        & \%Improv. & 97.74\% & 64.38\% & 2.58\% & 50.99\% & 67.93\% & 62.54\% \\
        \hline \hline
        \multirow{5}{*}{\makecell{INTER}} &
        ProtoBERT & 0.2312 & 0.0920 & 0.2713 & - & 0.2917 & 0.2216\\
        & NNShot & 0.2353 & 0.0634 & 0.2823 & - & 0.3280 & 0.2048 \\
        & ProML & 0.2456 & 0.0703 & 0.2650 & - & 0.2960 & 0.2192 \\
        & CONTaiNER & 0.2291 & 0.0687 & 0.3057 & - & 0.2681 & 0.2179 \\
        & CLLMFS & \textbf{0.4579} & \textbf{0.2392} & \textbf{0.4413} & - & \textbf{0.5128} & \textbf{0.4128} \\
        \hline
        & \%Improv. & 86.44\% & 160.0\% & 44.36\% & - & 56.34\% & 86.28\% \\
        \hline
    \end{tabular}
    }
    \label{tab:overall-performance-comparison}
\end{table*}

\begin{table}
\caption{Ablation Analysis.}
    \centering
    \resizebox{\linewidth}{!}{
    \begin{tabular}{l|c}
        \hline
        Modules & F1 Score\\
        \hline \hline
        LLAMA 2 + LoRA & 0.375 \\
        LLAMA 2 + LoRA + CL & 0.377 \\
        LLAMA 2 + LoRA + CL + Noise & 0.384 \\
        \hline
    \end{tabular}
    }
    \label{tab:ablation-analysis}
\end{table}

\begin{table}[h]
\caption{Impact of LoRA Module Parameter Combinations on F1-score.}
    \centering
    \resizebox{\linewidth}{!}{
    \begin{tabular}{ccccccc|c}
        \hline
        \(W_q\) & \(W_k\) & \(W_v\) & \(W_o\) & \(W_{in}\) & \(W_{out}\) & \(W_{wte}\) & F1-score \\
        \hline \hline
        $\checkmark$ & & & & & & & 0.281 \\
        & $\checkmark$ & & & & & & 0.283 \\
        & & $\checkmark$ & & & & & 0.350 \\
        $\checkmark$ & $\checkmark$ & & & & & & 0.329 \\
        $\checkmark$ & & $\checkmark$ & & & & & 0.370 \\
        $\checkmark$ & $\checkmark$ & $\checkmark$ & & & & & 0.367 \\
        $\checkmark$ & $\checkmark$ & $\checkmark$ & $\checkmark$ & & & & 0.360 \\
        $\checkmark$ & $\checkmark$ & $\checkmark$ & $\checkmark$ & $\checkmark$ & & & 0.368 \\
        $\checkmark$ & $\checkmark$ & $\checkmark$ & & $\checkmark$ & & & \textbf{0.375} \\
        $\checkmark$ & $\checkmark$ & $\checkmark$ & & $\checkmark$ & $\checkmark$ & & 0.372 \\
        $\checkmark$ & $\checkmark$ & $\checkmark$ & & $\checkmark$ & $\checkmark$ & $\checkmark$ & 0.373 \\
        $\checkmark$ & $\checkmark$ & $\checkmark$ & $\checkmark$ & $\checkmark$ & $\checkmark$ & $\checkmark$ & 0.352 \\
        \hline
    \end{tabular}
    }
    \label{tab:lora-parameter-impact}
\end{table}

\subsection{Dataset Description}
\label{ssec:dataset}

To assess the effectiveness of our method, we utilize 5 datasets spanning various domains: WNUT'17 \footnote{\url{https://huggingface.co/datasets/wnut_17}}, GUM \footnote{\url{https://gucorpling.org/gum/}}, I2B2 \footnote{\url{https://www.i2b2.org/NLP/DataSets/}}, OntoNotes \footnote{\url{https://www.ldc.upenn.edu/}}, and Conll2003 \footnote{\url{https://huggingface.co/datasets/conll2003}} which are publicly available and have been used in existing research \citep{HGB, KGNN-LS, KGCN, KGAT} to showcase diversity in terms of domain, scale, and sparsity. 

\begin{itemize}
    \item \textbf{WNUT'17} \citep{derczynski-etal-2017-results} is a collection of noisy user-generated text from social media platforms. This dataset contains annotations for 6 entity types, including 'corporation', 'creative-work', 'group', 'location', 'person', and 'product'.
    \item \textbf{GUM} \citep{zeldes2017gum} stands as a versatile, open-source multilayer resource, encompassing a spectrum of twelve text genres including narratives, interviews, news, instructions, and academic writing. It covers 11 entity types such as time, object, quantity, organization and other entities. 
    \item \textbf{I2B2} \citep{stubbs2015annotating} is annotated for Protected Health Information (PHI) and disease Risk Factors, serves as a critical resource within the medical domain. We specially focus on 6 entity recognition of PHI, like 'Patient ID', 'Hospital Location', 'Visit Date', 'Patient Profession', and 'Profession Contact'.
    \item \textbf{OntoNotes} \citep{weischedel2013ontonotes} is a large-scale, multilingual corpus that is collected from news, conversational telephone speech, weblogs and broadcast. This paper focuses on 18 entity types, including 'Geopolitical Entity', 'Organization', 'Person', 'Location', 'Money', 'Facility', 'Date', 'Ordinal', 'Quantity', 'Time', 'Nationalities, Religious or Political Groups', 'Cardinal', 'Percent', 'Event', 'Work of Art', 'Language', 'Law', and 'Product'.
    \item \textbf{CoNLL’03} \citep{tjong-kim-sang-de-meulder-2003-introduction} is also a benchmark dataset that focuses on 4 types of entities: persons, locations, organizations, and miscellaneous entities that do not belong to the previous three categories.
    
\end{itemize}

All of the above datasets use the N-way and 5-shot setting for training. For a fair comparison on those datasets, we split long sentences in some datasets into multiple shorter sentences to accommodate the input token limit of LLM, thus facilitating the extraction of text information by CLLMFS. We conducted tests on the WNUT'17, GUM, I2B2, OntoNotes, and CoNLL'03 datasets, utilizing approximately 1,200, 800, 750, 10,000, and 1,100 instances, respectively.


\subsection{Experimental Settings}
\label{ssec:es}

\subsubsection{Evaluation Metrics}
\label{sssec:baselines}

To compare our model with previous state-of-the-art (SOTA) models, we evaluate its performance by computing the micro-F1 score across the target domain.

\begin{itemize}
    \item \textbf{INTRA setting}: In traditional NER datasets such as WNUT’17, GUM, I2B2, OntoNotes, and CoNLL’03, distinct tag-set distributions are present. To address this, we generate multiple support sets by sampling from the original training set to train our model within the source domain. These support sets are subsequently employed for predictions on the original test set.
    
    \item \textbf{INTER (Cross Domain) setting}: In the cross-domain setting, our model is trained on the OntoNotes dataset, serving as the source domain, and subsequently tested on other datasets, constituting the target domain. The tag sets in different datasets are primarily determined by the dataset creators and often do not overlap. For instance, GUM is focused on social media terminology, whereas I2B2 is centered around medical terminology, leading to almost no overlap. In some rare cases where tag overlap occurs, the tags may still represent slightly different concepts (e.g., one dataset might use “place” to denote a neighborhood, while another might use “position” to refer to a city location). In this setting, the training and test set from OntoNotes is split into \textit{N-way K-shot} for training, and the test set consists of the original test sets from various domains, without utilizing their respective training sets.
\end{itemize}

\subsubsection{Baselines}
\label{sssec:baselines}

To evaluate CLLMFS's effectiveness, we compare it with several state-of-the-art Few-Shot NER models across various datasets and settings:

\begin{itemize}
\item \textbf{ProtoBERT} \citep{zhang2023less} simplifies Few-Shot NER using a span-based prototypical network with a pre-computable BERT encoder. It employs token embeddings to create entity prototypes and utilizes l-2 distance for efficient entity prediction during inference.
\item \textbf{NNShot} \citep{yang2020simple} adopts a novel token-level nearest neighbor classification approach, distinguishing itself from prototype-based methods by utilizing the proximity of similar samples in an embedding space.
\item \textbf{ProML} \citep{chen-etal-2023-prompt} introduces multiple prompt schemas to enrich label semantics and a novel architecture that synergistically integrates these prompts, advancing metric learning in Few-Shot NER.
\item \textbf{CONTaiNER} \citep{container} utilizes contrastive learning with Gaussian-distributed token embeddings to enhance Few-Shot NER. It focuses on optimizing generalized objectives to improve entity distinction without overfitting to specific domain attributes.
\end{itemize}

To ensure fair comparisons, we used the optimal parameters from each model’s respective code repositories. All models were trained and evaluated on the same datasets, with metrics averaged over five statistical runs for consistency.


\subsection{Performance Comparison}
\label{ssec:performance-comparison}

The performance comparison in Table \ref{tab:overall-performance-comparison} illustrates CLLMFS's superior effectiveness, achieving new state-of-the-art (SOTA) results. Across different datasets, CLLMFS shows substantial improvements over previous SOTA models, with average relative gains of 62.54\% and 86.28\% in micro F1 under the \textbf{INTRA} and \textbf{INTER} settings, respectively.

CLLMFS excels in various challenging scenarios, spanning both within-domain (INTRA) and cross-domain (INTER) NER tasks. Conventional baseline models, such as ProML, face difficulties in adapting to unseen text domains like GUM due to limited prompt design and methodological constraints. CONTaiNER, although effective in few-shot NER, struggles with substantial domain differences between source and target domains.

Despite these challenges, CLLMFS consistently outperforms SOTA models in both within-domain and cross-domain NER tasks, demonstrating robustness and adaptability. Moreover, CLLMFS effectively handles noisy data, as demonstrated in the WNUT'17 dataset, showcasing its suitability for real-world applications with varying data quality.


\subsection{Ablation Analysis}
\label{ssec:ablation-analysis}

We conducted ablation experiments to systematically investigate the impact of each constituent module on the performance of CLLMFS, which comprises three essential modules:  Low-Rank Adaptation (LoRA), Contrastive Learning (CL), and Uniform Gaussian Random Noise (Noise). Due to the complexity of computations involved in the LLAMA 2 model without utilizing LoRA technology, the computational resources available were insufficient to execute the model. Consequently, this experiment is excluded from consideration.

As depicted in Table \ref{tab:ablation-analysis}, we observed a clear trend of performance improvement with the inclusion of each additional module, which demonstrates the beneficial impact of incorporating Noise in conjunction with LoRA and CL, further bolstering the model's overall performance.


\subsection{Influence of LoRA module selections}
\label{ssec:influence-of-lora}

To enhance entity extraction tasks in low-resource settings, we systematically investigated the impact of LoRA module selections within the CLLMFS architecture. This architecture includes four weight matrices in the self-attention module \((W_q, W_k, W_v, W_o)\), two in the MLP module \((W_{in}, W_{out})\), and one for word token embeddings \((W_{wte})\). We applied LoRA to each weight matrix and explored the optimal configurations to maximize performance.

Using 5-fold cross-validation, we ensured the robustness of our findings, averaging performance over five iterations. The results, summarized in Table \ref{tab:lora-parameter-impact}, show that configurations involving LoRA on \(W_q\), \(W_k\), \(W_v\), and \(W_{in}\) consistently outperform others. However, adding LoRA to \(W_o\), \(W_{out}\) and \(W_{wte}\)does not consistently improve performance, as indicated by varying F1-scores across different configurations.

Overall, these findings underscore the importance of careful parameter tuning in optimizing the effectiveness of the LoRA module for few-shot NER tasks. The observed performance variations highlight the intricate interplay between different module parameters and their collective impact on model performance.


\section{Discussion}
\label{sec:discussion}

Our proposed CLLMFS framework achieves promising performance in NER task. Different from previous few-shot NER methods, CLLMFS fine-tunes the model and leverages LLM's capabilities by constructing entity SFT data from limited data, enhancing generalization for few-shot NER tasks. Different from previous meta learning methods, CLLMFS leverages abundant semantic information in LLMs, achieving consistent performance across target domains, even with limited source domain samples. Different from previous in-context learning methods, CLLMFS integrates SFT data for fine-tuning and introduces contrastive learning for FS-NER, enhancing boundary awareness and entity recognition accuracy. Please refer to the Appendix B for some study cases.


\section{Conclusion}
\label{sec:conclusion}

In this paper, we first propose CLLMFS by enhancing the large language model with contrastive learning for few-shot NER. Our method leverages the inherent knowledge within LLM and utilizes LoRA for supervised fine-tuning. By integrating contrastive learning, CLLMFS enhances LLM's ability of boundary awareness and entity extraction accuracy. Our approach has achieved state-of-the-art performance on multiple datasets with limited labeled data. The cross-domain experiment results confirm that the strong transfer capabilities of CLLMFS across different domains. In the future, we will concentrate on named entity recognition and extend our current work to relation extraction.




\newpage

\bibliography{mybibfile}

\end{document}



\begin{frontmatter}


\paperid{123} 


\title{CLLMFS: A Contrastive Learning enhanced Large Language Model Framework for Few-Shot Named Entity Recognition}

\author{
    Anonymous submission
}

\end{frontmatter}

\begin{appendices}

\section{Different entity types in SFT}

\subsection{Location}
\texttt{
\{
  "instruction": "Please extract the entity of location in the input sentence given below , the entity of place refers to the entity that represents the name or place of a specific location in the input sentence.",
  "input": "The hijackers told the crew they had grenades and other explosives and threatened to blow up the plane if they were not taken to London .",
  "output": "<im\_start> I can extract entities for you, the extracted entities are \text{<}\text{<}\text{<} London \text{>}\text{>}\text{>} <im\_end>"
\}
}

\subsection{Organization}
\texttt{
\{
  "instruction": "Please extract the entity of organization in the input sentence given below , the entity of organization refers to the entity that represents a specific organization in the input sentence.",
  "input": "There 's no problem whatsoever , \textbackslash " he told Reuters . He said the lifestyle associated with being Miss Universe could make routine exercise difficult .",
  "output": "<im\_start> I can extract entities for you, the extracted entities are \text{<}\text{<}\text{<} Reuters \text{>}\text{>}\text{>} <im\_end>"
\}
}

\subsection{Law}
\texttt{
\{
  "instruction": "Please extract the entity of Law in the input sentence given below, the entity of Law refers to the entity that represents a rule or system of rules recognized by a country or community in the input sentence.",
  "input": "but it seems to me unlikely on somebody that would be more of a strict constructionist on the constitution /.",
  "output": "<im\_start> I can extract entities for you, the extracted entities are \text{<}\text{<}\text{<} constitution \text{>}\text{>}\text{>} <im\_end>"
\}
}

\subsection{Work of Art}
\texttt{
\{
  "instruction": "Please extract the entity of Work of Art in the input sentence given below, the entity of Work of Art refers to the entity that represents a creative or artistic product or performance in the input sentence.",
  "input": "They were extremely disappointed with the sound on Autumn Moon , so they decided to go to Australia to redo it .",
  "output": "<im\_start> I can extract entities for you, the extracted entities are \text{<}\text{<}\text{<} Autumn Moon \text{>}\text{>}\text{>} <im\_end>"
\}
}

\subsection{Product}
\texttt{
\{
  "instruction": "Please extract the entity of product in the input sentence given below , the entity of product refers to the entity that represents the name or category of a specific product in the input sentence.",
  "input": " New \#VideoRecorder \#7 : Flip MinoHD Video Camera \--\-- 8 GB , 2 Hour Record Time : Flip MinoHD Video Camera \--\-- ... http://dlvr.it/5P5kV \#cheaptweet",
  "output": "<im\_start> I can extract entities for you, the extracted entities are \text{<}\text{<}\text{<} Flip MinoHD \text{>}\text{>}\text{>} \text{<}\text{<}\text{<} Flip MinoHD \text{>}\text{>}\text{>} <im\_end>"
\}
}

\subsection{Corporation}
\texttt{
\{
  "instruction": "Please extract the entity of corporation in the input sentence given below , the entity of corporation refers to the entity that represents the name or organization of a specific corporation in the input sentence.",
  "input": " Jackie joins Twitter 50 years after JFK campaign [ AP : US News ] http://bzbx.us/uEf \#50 \#Campaign via buzzbox . com",
  "output": "<im\_start> I can extract entities for you, the extracted entities are \text{<}\text{<}\text{<} Twitter \text{>}\text{>}\text{>} <im\_end>"
\}
}

\section{Study Cases}

\setcounter{table}{0}
\renewcommand{\thetable}{A\arabic{table}}

\subsection{Few-shot cases}

The term "None" in Table~\ref{tab:case1} indicates that the model produced no output. The results demonstrate that our approach exhibits superior semantic understanding in low-resource NER tasks, effectively identifying entities such as 'Andrew' and 'Little'. CLLMFS demonstrates the ability to extract the complete person name entity from the sentence, while other methods are best at extracting features 'Andrew' . Consequently, we assert that CLLMFS offers superior semantic understanding and is less susceptible to being influenced solely by the inherent meaning of the word 'Little'. This advantage can be attributed to the incorporation of LLAMA2 in our methodology.

In the case of Table~\ref{tab:case2}, it can be observed that NNshot and ProtoBERT methods exhibit limited performance in recognizing specialized noun entities, indicating their constrained generalization capability and susceptibility to neighboring characters' influence. CONTaiNER and ProML successfully identify the target entities but fail to capture the semantics of specialized nouns, resulting in the extraction of irrelevant entities. In contrast, our method, based on a large-model architecture, accurately identifies specialized noun entities like 'Trump administration' without extracting any erroneous entities.

\begin{table*}
\caption{Case A1.}
    \centering
    \resizebox{\linewidth}{!}{
    \renewcommand{\arraystretch}{1.5} 
    \begin{tabular}{|c|>{\centering\arraybackslash}p{0.2\linewidth}|>{\centering\arraybackslash}p{0.2\linewidth}|>{\centering\arraybackslash}p{0.2\linewidth}|>{\centering\arraybackslash}p{0.2\linewidth}|>{\centering\arraybackslash}p{0.2\linewidth}|}
        \hline
        \textbf{Input} & \multicolumn{5}{c|}{\thead{True, but I imagine it would be a lot lower and as I pointed out to Andrew Little \\ would be cheaper than [ eliminating fees.}} \\
        \hline
        \textbf{label} & \multicolumn{5}{c|}{\thead{'Andrew', 'Little'}} \\
        \hline \hline
        \textbf{Model} & ProtoBERT & NNShot & ProML & CONTaiNER & CLLMFS \\
        \hline
        \textbf{Output} & 'Andrew' & 'but', 'it', 'be', 'as', 'out' & 'Andrew' & None & 'Andrew', 'Little' \\
        \hline
    \end{tabular}
    }
    \label{tab:case1}
\end{table*}

\begin{table*}
\caption{Case A2.}
    \centering
    \resizebox{\linewidth}{!}{
    \renewcommand{\arraystretch}{1.5} 
    \begin{tabular}{|c|>{\centering\arraybackslash}p{0.2\linewidth}|>{\centering\arraybackslash}p{0.2\linewidth}|>{\centering\arraybackslash}p{0.2\linewidth}|>{\centering\arraybackslash}p{0.2\linewidth}|>{\centering\arraybackslash}p{0.2\linewidth}|}
        \hline
        \textbf{Input} & \multicolumn{5}{c|}{\thead{/ . Trump administration rolls back protections for people in default on student loans https://t.co/YPe2DRsjLe
        }} \\
        \hline
        \textbf{label} & \multicolumn{5}{c|}{\thead{'Trump', 'administration'}} \\
        \hline \hline
        \textbf{Model} & ProtoBERT & NNShot & ProML & CONTaiNER & CLLMFS \\
        \hline
        \textbf{Output} & '/', 'default' & '.', 'Trump', 'back', 'protections', 'for', 'default', 'on', 'loans' & 'Trump', 'protections', 'default', 'loans' & 'administration', 'loans' & 'Trump', 'administration' \\
        \hline
    \end{tabular}
    }
    \label{tab:case2}
\end{table*}

\subsection{Contrastive learning cases}

In Table~\ref{tab:case3}, Lack of contrastive learning in the model results in poor boundary perception, thereby leading to inadequate semantic comprehension and increased likelihood of over-extraction or under-extraction in low-resource entity extraction tasks.

In the case of Table~\ref{tab:case4}, it is evident that without contrastive learning, not only was the entity 'the world' extracted, but also other entities such as 'some great musicians'. By introducing contrastive learning as an auxiliary task, CLLMFS have effectively improved the internal representations of the large model, resulting in more precise entity extraction.

\begin{table*}
\caption{Case A3.}
    \centering
    \resizebox{\linewidth}{!}{
    \renewcommand{\arraystretch}{1.5} 
    \begin{tabular}{|c|>{\centering\arraybackslash}p{0.4\linewidth}|>{\centering\arraybackslash}p{0.4\linewidth}|}
        \hline
        \textbf{Input} & \multicolumn{2}{c|}{\thead{topless sunbathing is common by western women at many tourist beaches .}} \\
        \hline
        \textbf{label} & \multicolumn{2}{c|}{\thead{'many tourist beaches'}} \\
        \hline \hline
        \textbf{Model} & CLLMFS without CL & CLLMFS \\
        \hline
        \textbf{Output} & 'tourist beaches' & 'many tourist beaches' \\
        \hline
    \end{tabular}
    }
    \label{tab:case3}
\end{table*}

\begin{table*}
\caption{Case A4.}
    \centering
    \resizebox{\linewidth}{!}{
    \renewcommand{\arraystretch}{1.5} 
    \begin{tabular}{|c|>{\centering\arraybackslash}p{0.4\linewidth}|>{\centering\arraybackslash}p{0.4\linewidth}|}
        \hline
        \textbf{Input} & \multicolumn{2}{c|}{\thead{i worked with some great musicians and performed around the world .
        }} \\
        \hline
        \textbf{label} & \multicolumn{2}{c|}{\thead{'the world'}} \\
        \hline \hline
        \textbf{Model} & CLLMFS without CL & CLLMFS \\
        \hline
        \textbf{Output} & 'the world', 'some great musicians' & 'the world' \\
        \hline
    \end{tabular}
    }
    \label{tab:case4}
\end{table*}

\subsection{Transfer learning cases}

The examples below are all trained on the OntoNotes source data and evaluated on the I2B2, WNUT'17, GUM, and CoNLL'03 datasets, respectively. Across these datasets, our approach demonstrates robust cross-domain NER capabilities.

In the medical domain dataset I2B2 (Table~\ref{tab:case5}), our method excels in identifying relevant entities, while CONTaiNER and ProML, although capable of extraction, introduce too many erroneous entities. In the social media dataset WNUT'17 (Table~\ref{tab:case6}), our method effectively identifies location entities, whereas ProML successfully recognizes them but also extracts other entity information. In the GUM dataset (Table~\ref{tab:case7}), our method uniquely identifies relevant entities correctly. In the news dataset CoNLL'03 (Table~\ref{tab:case8}), only our method and NNShot can accurately identify the journalistic terminology 'reuters'. In summary, this demonstrates the strong cross-domain entity extraction capability of CLLMFS.

\begin{table*}
\caption{Case A5.}
    \centering
    \resizebox{\linewidth}{!}{
    \renewcommand{\arraystretch}{1.5} 
    \begin{tabular}{|c|>{\centering\arraybackslash}p{0.2\linewidth}|>{\centering\arraybackslash}p{0.2\linewidth}|>{\centering\arraybackslash}p{0.2\linewidth}|>{\centering\arraybackslash}p{0.2\linewidth}|>{\centering\arraybackslash}p{0.2\linewidth}|}
        \hline
        \textbf{Input} & \multicolumn{5}{c|}{\thead{she came from turlock 10 years ago to pecos to be with her daughter who is a surgeon in the area}} \\
        \hline
        \textbf{label} & \multicolumn{5}{c|}{\thead{'surgeon'}} \\
        \hline \hline
        \textbf{Model} & ProtoBERT & NNShot & ProML & CONTaiNER & CLLMFS \\
        \hline
        \textbf{Output} & 'she', 'came', 'ago', 'be', 'is' & None & 'pecos', 'her', 'daughter', 'who', 'is', 'a', 'surgeon', 'in', 'area' & 'pecos', 'with', 'her', 'daughter', 'who', 'a', 'surgeon', 'in', 'area' & 'surgeon' \\
        \hline
    \end{tabular}
    }
    \label{tab:case5}
\end{table*}

\begin{table*}
\caption{Case A6.}
    \centering
    \resizebox{\linewidth}{!}{
    \renewcommand{\arraystretch}{1.5} 
    \begin{tabular}{|c|>{\centering\arraybackslash}p{0.2\linewidth}|>{\centering\arraybackslash}p{0.2\linewidth}|>{\centering\arraybackslash}p{0.2\linewidth}|>{\centering\arraybackslash}p{0.2\linewidth}|>{\centering\arraybackslash}p{0.2\linewidth}|}
        \hline
        \textbf{Input} & \multicolumn{5}{c|}{\thead{i looked this up yesterday actually ; the average household income for auckland in 2016 was \$ 104 k .}} \\
        \hline
        \textbf{label} & \multicolumn{5}{c|}{\thead{'auckland'}} \\
        \hline \hline
        \textbf{Model} & ProtoBERT & NNShot & ProML & CONTaiNER & CLLMFS \\
        \hline
        \textbf{Output} & 'I', 'household', 'in', '2016' & 'household', '2016' & 'this', 'Auckland', '104', 'k', '.' & None & 'auckland' \\
        \hline
    \end{tabular}
    }
    \label{tab:case6}
\end{table*}

\begin{table*}
\caption{Case A7.}
    \centering
    \resizebox{\linewidth}{!}{
    \renewcommand{\arraystretch}{1.5} 
    \begin{tabular}{|c|>{\centering\arraybackslash}p{0.2\linewidth}|>{\centering\arraybackslash}p{0.2\linewidth}|>{\centering\arraybackslash}p{0.2\linewidth}|>{\centering\arraybackslash}p{0.2\linewidth}|>{\centering\arraybackslash}p{0.2\linewidth}|}
        \hline
        \textbf{Input} & \multicolumn{5}{c|}{\thead{most travellers will arrive in York by road ( car or bus ) or rail from other parts of the uk or an airport .}} \\
        \hline
        \textbf{label} & \multicolumn{5}{c|}{\thead{'most travellers'}} \\
        \hline \hline
        \textbf{Model} & ProtoBERT & NNShot & ProML & CONTaiNER & CLLMFS \\
        \hline
        \textbf{Output} & 'in', 'York', 'by', 'car', 'parts', 'of', 'an' & 'York', 'car', 'other' & 'bus', 'rail', 'from', 'other', 'parts', 'of', 'or' & 'York', 'road', 'rail', 'from', 'other', 'parts' & 'most travellers' \\
        \hline
    \end{tabular}
    }
    \label{tab:case7}
\end{table*}

\begin{table*}
\caption{Case A8.}
    \centering
    \resizebox{\linewidth}{!}{
    \renewcommand{\arraystretch}{1.5} 
    \begin{tabular}{|c|>{\centering\arraybackslash}p{0.2\linewidth}|>{\centering\arraybackslash}p{0.2\linewidth}|>{\centering\arraybackslash}p{0.2\linewidth}|>{\centering\arraybackslash}p{0.2\linewidth}|>{\centering\arraybackslash}p{0.2\linewidth}|}
        \hline
        \textbf{Input} & \multicolumn{5}{c|}{\thead{he told reuters he had needed to speak to her before she left wellington later on friday .}} \\
        \hline
        \textbf{label} & \multicolumn{5}{c|}{\thead{'reuters'}} \\
        \hline \hline
        \textbf{Model} & ProtoBERT & NNShot & ProML & CONTaiNER & CLLMFS \\
        \hline
        \textbf{Output} & 'reuters', 'wellington' & 'reuters' & None & None & 'reuters' \\
        \hline
    \end{tabular}
    }
    \label{tab:case8}
\end{table*}

\end{appendices}